\documentclass{kais}
\usepackage{wrapfig, epsf}
\usepackage[dcucite]{harvard}

\usepackage{enumerate} 
\usepackage{graphicx} 
\usepackage{tikz} 
\usepackage[noend]{algorithmic} 
\usepackage{algorithm} 
\usepackage{amsmath, amsthm, amssymb} 
\usepackage{soul}

\usepackage{subfigure}

\received{xxx}
\revised{xxx}
\accepted{xxx}

\pubyear{2012}
\pagerange{\pageref{firstpage}--\pageref{lastpage}}
\volume{xxx}

\begin{document}
\label{firstpage}

\title{Crowd Labeling: A Survey}

\author[J. Muhammadi, H. R. Rabiee and A. Hosseini]{Jafar Muhammadi, Hamid R. Rabiee, Abbas Hosseini\\ 
Department of Computer Engineering, Sharif University of Technology, Tehran, Iran. }

\maketitle

\begin{abstract}
Recently, there has been a burst in the number of research projects on human computation via crowdsourcing. Multiple choice (or labeling) questions could be referred to as a common type of problem which is solved by this approach. As an application, crowd labeling is applied to find true labels for large machine learning datasets. 
Since crowds are not necessarily experts, the labels they provide are rather noisy and erroneous. This challenge is usually resolved by collecting multiple labels for each sample, and then aggregating them to estimate the true label. Although the mechanism leads to high-quality labels, it is not actually cost-effective. As a result, efforts are currently made to maximize the accuracy in estimating true labels, while fixing the number of acquired labels. 

This paper surveys methods to aggregate redundant crowd labels in order to estimate unknown true labels. It presents a unified statistical latent model where the differences among popular methods in the field correspond to different choices for the parameters of the model. Afterwards, algorithms to make inference on these models will be surveyed. Moreover, adaptive methods which iteratively collect labels based on the previously collected labels and estimated models will be discussed. In addition, this paper compares the distinguished methods, and provides guidelines for future work required to address the current open issues. 
\end{abstract}


\section{Introduction} 
Crowdsourcing has become a popular field of research during recent years. In crowdsourcing, a group of people are asked to contribute in performing a task that cannot be individually done with the same ease \cite{crowdsourcing}. ``Wikipedia", for instance, is a well recognized crowdsourcing system, in which thousands of Internet users participate in the creation of the largest worldwide encyclopaedia. Crowdsourcing brings about a large number of applications, ranging from funding (e.g. KickStarter which funds creative projects) and forecasting (e.g. Threadless which estimates the success rate of T-shirt designs in the market) to organization (e.g. Digg which organizes Internet links) and human computation. 

Human computation is defined as harnessing human intelligence to solve computational problems beyond the scope of existing Artificial Intelligence (AI) algorithms \cite{HCBook}. One of the main reasons that AI has not yet achieved all human capabilities is that it lacks common sense knowledge, whatever facts a human knows (captures, saves, and uses) \cite{Waltz}. In 1992, the problem was originally proposed by Marvin Minski in the context of slow progress in processing natural languages \cite{AIFMinsky}. He stated that computers do not have access to the meanings of the words and objects, as humans do. With the ``ROPE", as an example, someone can pull something, but not push it, or he can wrap something, but not eat it, etc. Even a child can describe more than a hundred applications of a rope, or any other object or words, in a few minutes. But, a computer can not do so. Creating common sense knowledge-base is a very demanding task, because (i) a huge amount of information must be captured, (ii) there is no proper method to represent the knowledge, (iii) updating the facts is very difficult, and (iv) there is a lack of efficient methods to use and make inference about that knowledge \cite{McCarthyHLAI,MOM}. 

When large-scale dramatic problems have to be overcome within a limited budget, human intelligence can be harnessed using crowdsourcing. For example, it is common place to find true labels for large Machine Learning (ML) datasets through human computation. It means that human computation using crowdsourcing is becoming the spurt of AI and ML research. As the following section show,  most human computation applications can be described as a set of labeling problems. Therefore, solving labeling problems using crowdsourcing, or briefly ``crowd labeling'', is becoming a hot research topic. Although crowd labeling is an important and growing field of inquiry, there has not been any notable survey concerning the area. Therefore, we believe that such surveys are not only helpful, but necessary as well. 

This paper is aimed at introducing crowd labeling and its complexities, and surveying the literature that attempt to overcome those complexities. Then, it presents a unified statistical latent model where the differences among popular methods in the field correspond to different choices for the parameters of the model. The paper has both practical and theoretical contributions. It systematically exposes the main aspects of crowd labeling methods. Moreover, it compares the performance of different methods using comprehensive experiments. 

The reminder of this paper is organized as follows: In section 2, we introduce crowd labeling. Section 3 is where the target problem of the paper will be formally defined. Section 4 presents the assumed latent models in label aggregation process. Section 5 intends to survey the inference algorithms of latent models. Section 6 describes adaptive methods. Section 7 includes experimental results. Finally, concluding remarks, open issues, and guidelines for future work are presented in the last section. 

\section{Crowd Labeling}
There are some constraints in using crowdsourcing for human computation. Since the participants are not experts, the problems are supposed to be normally small, simple, and well-formed. In addition, due to human error and bias, it is mandatory to ensure that collected responses are adequately reliable. Dividing the main problem into several micro-problems is a helpful way to resolve the first challenge. And, there must be a quality control mechanism to overcome the second one. 

Labeling (or multiple-choice) questions are the simplest well-known type of micro-problems in which the answering process is to simply choose one of the provided choices. Most of other more complex types of problems are convertible to this type \cite{FrankHall2001,AttImgRank,Soylent,Little}. And finally, there are efficient quality assuring mechanisms addressing labeling questions. Many applications have used labeling micro-problems to utilize humans' power. For example, evaluating the accuracy of the results returned by search engines \cite{crowdsearch}; shortening and proofreading of documents \cite{Soylent}; quantifying the ability of native listeners to perform speaker recognition \cite{SpeakerRec}; classifying galaxies from the Sloan Digital Sky Survey \cite{GalaxyZoo}; semantic similarity detection in maps \cite{OSM}; and, solving object recognition problems using visual 20 question games \cite{Hloop}. 

In order to remove noise, bias, and errors, the solutions provided by humans should be validated. In multiple checking mechanisms, the solution of each problem is requested from multiple labelers. Then, collected responses are aggregated (e.g. using majority voting) to find the final solution of that problem. In ``collaborative tasks'', users build on or evaluate each other's answers \cite{collorgs}. For example, in ``find-fix-verify'' mechanism, some users specify the problems (``find'' operation), other users solve them (``fix'' operation), and some others are supposed to confirm that the provided solutions are valid (``verify'' operation) \cite{Turkit,Soylent}. And, in Game With A Purpose (GWAP) scenarios, context-dependent proprietary procedures are used for quality assurance (See the next paragraph for more information). Multiple checking is the most frequently mechanism used for quality assurance in labeling problems. 

Another issue in crowd labeling is finding and utilizing the humans. There is a large number of micro-problems in each application pertinent to crowd labeling. A large number of humans are required to solve these problems. Let's consider Content Based Image Retrieval (CBIR) using a game for example. ESP method \cite{ESP} involves a large number of humans in an interesting game which subsequently results in some tagged images. In this game, the system randomly pairs two online users who neither know, nor communicate with each other. An Image is displayed to the players, and during a specified period of time, the players independently guess the image content by presenting text tags. Considering the constraint that they are not allowed to use the words presented in a taboo list, the players win the game only if one of them presents a tag which has been already presented by the other (the taboo list is provided by the system in order to exclude the obvious tags or the tags previously obtained for that image in other games). The resulting tag is then used as a new annotation for that image. In this approach, a huge number of game players are required to collect tags for images in a real-world multimedia search engine database.

Since GWAP is not applicable to any given problem, labor marketplaces are used in most cases. Such marketplaces require a marginal cost to provide human responses to micro-problems. Amazon Mechanical Turk (MTurk)\footnote{http://mturk.com} is a famous crowdsourcing marketplace. It has been illustrated that the quality of the majority vote of multiple responses collected by MTurk is at least as good as that of answers provided by individual experts \cite{CheapFast}. CrowdFlower\footnote{http://crowdflower.com}, CastingWords\footnote{http://castingwords.com}, CrowdSpring\footnote{http://crowdspring.com}, Microworkers\footnote{http://microworkers.com}, and MobileWorks\footnote{http://mobileworks.com} are other samples of labor marketplaces. 

Push and pull are two approaches for task routing in crowdsourcing platforms. In the former, the system determines which tasks should be assigned to each user. In the latter, the system does not enforce explicitly its decisions, but rather does task routing implicitly by setting up the environment for workers \cite{HCBook}. Since game players have no choice in deciding which questions they are asked, the push approach is suitable for GWAP. There has also been some research projects on push approaches in labor marketplaces \cite{AdaptiveTA1,AdaptiveTA2,AdaptiveTA3}. However, none of the famous marketplaces use this approach. In typical marketplaces, the problems are designed in a way which maximizes the probability of being selected by groups of users. This requires being aware of the criteria considered by the users in selecting the problems. Samples of such criteria are: the time of importing the problems to the pool, the expiration time set for each problem, and the reward assigned to each problem. The proper values relevant to such criteria can be estimated through user behavior assessment. \cite{FinIncentives,CondBehaviour,Soylent,SearchHCMarket,Zhu2010,QualityManagement,AnnotRanking}. For example, in order to assess the role of reward in MTurk for a set of problems, \cite{Soylent} imported several problems with various amount of rewards to the system, and measured the performance of the system in each case.
They found that decreasing the amount of the reward does not affect the quality of the solutions; while it increases the waiting time. 

\section{Problem Configuration} 
The target problem of the paper is using a crowdsourcing marketplace (e.g. MTurk) to find high-quality labels for a large set of samples (e.g. providing labels for an ML dataset). The following assumptions are taken into considerations: (i) Budget is limited, (ii) A multiple checking mechanism is used for quality assurance, and iii) The pull approach is used for task routing (there is no task assignment mechanism). 
In addition, there is no limitation on the number of classes. However, in order to avoid complexities in tracking the subject, in some cases binary labeling will be considered. In most cases, the resulting algorithms and equations of binary labeling problems are extensible to general labeling problems. 

Let's assume that there are samples $\boldsymbol{X}=\{x_i\}_{i=1:N}$ with unknown true labels $\boldsymbol{Y}=\{y_i\}_{i=1:N}$. Users of the system are denoted by $\boldsymbol{U}=\{u_j\}_{j=1:R}$. 
The labels which are provided by users for samples are stored in $\boldsymbol{A}=\{A_{ij}\}_{i=1:N, j=1:R}$. The provided labels for sample $x_i$, i.e. the $i$th row of $\boldsymbol{A}$ is denoted by $\boldsymbol{A}_i$. The unseen labels in the matrix $\boldsymbol{A}$ are shown by $0$. The total number of all collected labels for all samples (the budget) is limited, i.e. $\sum_{i=1}^N{\sum_{j=1}^R{I(A_{ij}\neq0)}}=B$, where $I$ is the indicator function. The goal is to find $\boldsymbol{\hat{Y}}=\{\hat{y}_i\}_{i=1:N}$ which maximizes $P(\boldsymbol{\hat{Y}}=\boldsymbol{Y}|\boldsymbol{A})$. 

The base method for integrating the collected labels and estimating the $\hat{y_i}$s is majority voting. Label aggregation has been recently done on the basis of more efficient methods which are surveyed in the following sections. 

\section{Latent Models}
\label{Sec:LatentModels}
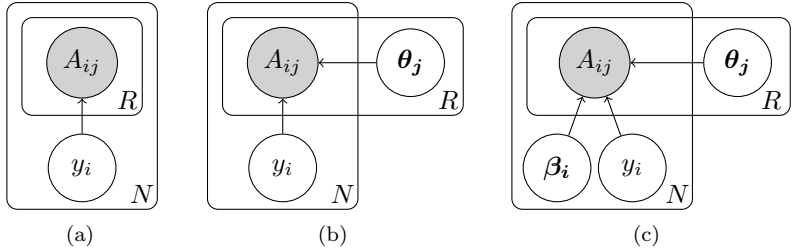
\begin{figure}
\centering

\subfigure[]{
\begin{tikzpicture} 
\draw[rounded corners=4pt] (0, 0) rectangle (2, -2.75);
\draw[rounded corners=4pt] (0.2, -0.2) rectangle (1.8, -1.5);
\node at (1.6, -1.3) {$R$};
\node at (1.8, -2.55) {$N$};

\node [circle, draw, fill=gray!35, minimum size=0.9cm] (a1) at (1, -0.8){$A_{ij}$};
\node [circle, draw, minimum size=0.9cm] (b1) at (1, -2.2){$y_i$};
\draw [->] (b1) -- (a1); 
\end{tikzpicture} 
\label{MVLM}
}
~
\subfigure[]{
\begin{tikzpicture} 
\draw[rounded corners=4pt] (0, 0) rectangle (2, -2.75);
\draw[rounded corners=4pt] (0.2, -0.2) rectangle (3.4, -1.5);
\node at (3.2, -1.3) {$R$};
\node at (1.8, -2.55) {$N$};

\node [circle, draw, fill=gray!35, minimum size=0.9cm] (a1) at (1, -0.8){$A_{ij}$};
\node [circle, draw, minimum size=0.9cm] (b1) at (1, -2.2){$y_i$};
\node [circle, draw, minimum size=0.9cm] (c1) at (2.7, -0.8){$\boldsymbol{\theta_j}$};
\draw [->] (b1) -- (a1); 
\draw [->] (c1) -- (a1); 
\end{tikzpicture} 
\label{DSLM}
}
~
\subfigure[]{
\begin{tikzpicture} 
\draw[rounded corners=4pt] (0, 0) rectangle (2.4, -2.75);
\draw[rounded corners=4pt] (0.2, -.2) rectangle (3.7, -1.5);
\node at (3.45, -1.3) {$R$};
\node at (2.2, -2.55) {$N$};

\node [circle, draw, fill=gray!35, minimum size=0.9cm] (a1) at (1.1, -0.8){$A_{ij}$};
\node [circle, draw, minimum size=0.9cm] (b1) at (.6, -2.2){$\boldsymbol{\beta_i}$};
\node [circle, draw, minimum size=0.9cm] (b2) at (1.6, -2.2){$y_i$};
\node [circle, draw, minimum size=0.9cm] (c1) at (3, -0.8){$\boldsymbol{\theta_j}$};

\foreach \from/\to in {b1/a1, b2/a1, c1/a1} 
\draw [->] (\from) -- (\to); 
\end{tikzpicture} 
\label{DARELM}
}
\caption{The graphical models of latent models in crowd labeling, from simple (a) to complex (c). }
\label{GraphicalModels}
\end{figure}

To classify each sample, the labels provided by multiple individuals are aggregated. Majority voting is the base method for label aggregation. More advanced methods are capable of taking into account factors such as user capabilities and problem difficulties. These methods build statistical latent models that determine how the labels are generated by labelers. 
The graphical models of these latent models are shown in Fig. \ref{GraphicalModels}. In these graphs, observed variables are depicted using shaded nodes and blank nodes represent latent variables. Moreover, arrows are used to represent the dependency among random variables. The plate structure is used to act as a for loop to represent repetition. As it can be seen, there are totally $N$ samples with unkown true labels $y_i$ and there are $R$ responses to each of them provided by users.

\begin{figure}\center{\includegraphics[scale=0.7]{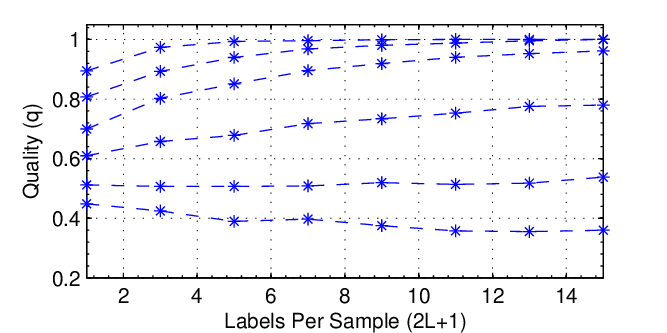}} 
\caption{The quality of estimated labels ($q$), based on the number ($2L+1$) and the quality ($\alpha$) of collected labels. The value of $\alpha$ is equal to $q$ where $L$ is $0$.} 
\label{MV-NQoU}
\end{figure}

Fig. \ref{MVLM} represents the latent model for majority voting. According to this model, it is assumed that the provided labels for a sample  depend only on the true label. That is, users provide correct answers to questions with a common known probability $\alpha$ (the users' accuracy in answering questions is $\alpha$). In this method, 
\begin{eqnarray}
\label{MVEq}
P(y_i=y|\boldsymbol{A}_i) = \frac{P(\boldsymbol{A}_i|y_i=y)P(y_i=y)}{P(\boldsymbol{A}_i)} \propto \prod_{j=1}^R{P(A_{ij}|y_i=y)}
\end{eqnarray}
Since $log(.)$ is a monotonically increasing function, it may be shown that 
\begin{eqnarray}
P(y_i=y|\boldsymbol{A}_i) \propto \left(\alpha\sum_{j=1}^R {I(A_{ij}=y)}\right)\left( (1-\alpha)\sum_{j=1}^R {I(A_{ij} \neq y)}\right)
\end{eqnarray}
Considering $\alpha \rightarrow 1$, the familiar equation of majority voting appears. This indicates that $P(y_i=y|\boldsymbol{A}_i)$ is merely depends on $\alpha$ and the number of $y$ votes. To realize the role of $\alpha$ and the number of participants in majority voting, suppose that ${z_1, \ldots, z_{2L+1}}$ denote the collected labels for sample $x_i$, in a binary labeling problem. Using (\ref{MVEq}), the correctness probability of the aggregated solutions is: 
\begin{equation} 
\label{MV}
q=P(\hat{y_i}=y_i)=\sum_{k=0}^{L}{\begin{pmatrix} 2L+1\\k \end{pmatrix}\alpha ^{2L+1-k}(1-\alpha)^{k}} 
\end{equation} 
where $q$ indicates the probability that the majority vote is the correct answer, i.e. the probability that more than $L$ labelers propose correct labels. Moreover, $\hat{y_i}$ is the estimated label by majority voting for $x_i$, and $k$ indicates the possible number of incorrect answers.
According to (\ref{MV}), $q$ is bigger than $p$, if $\alpha >0. 5$. Furthermore, in the case that $\alpha >0. 5$, as $L$ increases, there is a growth in $q$, but a decrease in the rate of changes. For example, increasing the number of labelers leads to more significant results when $\alpha =0.7$, compared with the case where $\alpha =0. 9$ \cite{GALabel}. 
Fig. \ref{MV-NQoU} shows the value of $q$ for different values of $\alpha$, based on the number of collected labels per sample. 

Majority voting is an intuitive and simple method. Moreover, its low computational cost, makes it a frequently used method for label aggregation in crowd labeling. However, as it is depicted in Fig. \ref{MV-NQoU}, when $\alpha$ is small, a large number of labels are required to estimate the true labels accurately. Furthermore, it has some simplifying assumptions which are not true in real world. In most labor marketplaces, there are various kinds of people with different accuracies in answering, but majority voting does not consider the quality of responses. As an example, there are spammers that try to maximize their income by answering questions as fast as possible, without caring about the true answers. 

More advanced methods, consider user capabilities. As shown in Fig. \ref{DSLM}, these methods model the capability of user $u_j$ by a (set of) random variable(s) $\boldsymbol{\theta_j}$. Different methods have been proposed for modeling the user capabilities. One of the simplest methods is to model each user's capability by a single random variable which corresponds to the user's accuracy \cite{txteagle,VarInf,SVD1,SVD2}. That is, the probability that user $u_j$ answers a problem correctly is $\boldsymbol{\theta_j}=\alpha_j$, where $\alpha_j \in \left[0, 1\right]$. $\alpha_j=1$ indicates that the user $u_j$ is an expert, $\alpha_j\simeq0.5$ means that the user $u_j$ is a spammer, and $\alpha_j<0.5$ shows that $u_j$ is an adversary user. These methods simultaneously estimate the true labels and user's capabilities by making inference on the model using the collected labels. 
The model reduces to majority voting if $\alpha_j$ is considered equal for all users.
The likelihood that the label provided by user $u_j$ is the correct label is a Bernoulli distribution, that is:
\begin{eqnarray}
p(A_{ij}|\alpha_j,y_i)=\alpha_j^{I(A_{ij}=y_i)}(1-\alpha_j)^{I(A_{ij}\neq y_i)}
\end{eqnarray} 
In order to make the inference tractable, a Beta distribution, $\beta(a,b)$, on $\alpha_j$ is usually chosen as the prior, which is conjugate to the labels' likelihood, i.e. Bernoulli distribution \cite{VarInf}. The hyperparameters $a$ and $b$ represent the prior knowledge about the accuracy of users. Actually, the $\beta(a,b)$ prior over $\alpha_j$ can be interpreted as observing $a+b$ pseudo-labels from $u_j$ which $a$ of them are correct. It can be shown that  $E[\alpha|a,b]=\frac{a}{a+b}$ and hence, if our prior knowledge indicates that the most of the users are spammers,  we should set $a=b$ and for the case that accuracies of most users is more than $0.5$, $a$ should be set greater than $b$. Moreover, $a+b$ indicates the uncertainty in our prior knowledge about user accuracies. Therefore, in the case that we don't have any prior knowledge about the accuracy of the users, choosing $a=b=\epsilon$  where $\epsilon \rightarrow 0^+$ is appropriate. This is equivalent to Haldane distribution, $\frac{1}{\alpha_j(1-\alpha_j)}$ \cite{Haldane}.

When users have different accuracies, but a user can answer questions of different categories with almost the same accuracy, the method leads to significant improvements. In addition, since the number of latent variables is small, a few observations are sufficient to efficiently estimate the latent variables. However, the method fails when the expertises of users on different categories varies a lot.

A common method to model the user expertises on different categories of problems, is to model the capability of user $u_j$ by a confusion matrix, $\boldsymbol{\pi^j}$ \cite{EM1979}. The element $\pi_{kl}^j$ of the matrix measures the probability that the user $u_j$ may provide label $l$ for a given problem whose true label is $k$. In binary labeling problems, confusion matrix reduces to sensitivity and specificity measurements, which indicate the proportion of actual positive and negative samples that are accurately recognized, respectively. In this method, Beta and its multivariate generalization, Dirichlet, are also considered as prior distribution over $\boldsymbol{\pi^j}$ \cite{VarInf,CrowLearn,Truelabel}.
This method leads to very good results in applications that include problems with almost the same level of difficulties. Hence, this method is very popular and is followed in many other researches \cite{venus,QualityManagement,PMFC2}. 
However, it should be noted that the number of latent variables in this method is more than the previous method and hence, it requires more user labels to estimate the capability of users, i.e., its data complexity is higher than the previous method.

More complex methods consider problem properties beside user capabilities. There is evidence that it is useful for some applications to take the difficulty level of problems into account. For example, the experiments on different categories of problems in \cite{Workforce} depict that using the same users for all problem categories leads to different results. In addition, in some categories the majority vote is better than the answers of the best user in the system, while in some others the opposite is true. Furthermore, in some problems the diversity of user's answers is very high, while in some others it is low \cite{ClassCrowd}. 
The general latent model for such methods is depicted in Fig. \ref{DARELM}. As shown in the figure, in these methods, the difficulty level of problem $x_i$ is modeled by a (set of) random variable(s) $\boldsymbol{\beta_i}$. Different methods have been proposed for modeling the user capabilities and problem difficulty levels. In GLAD method \cite{GLADPaper}, the difficulty level of problem $x_i$ is modeled by single random variable $\beta_i \in [0, \infty)$ and the capability of user $u_j$ is modeled by $r_j \in (-\infty, +\infty)$. In addition, the label generation model is assumed as the following logistic model,
\begin{equation}
P(A_{ij}=y_i|r_j, \beta_i)=\frac{1}{1+e^{-r_j \beta_i}}
\end{equation}
According to this model, the probability of providing correct label for sample $x_i$ by user $u_j$ increases as $1/\beta_i$ decreases and it approaches $0.5$ as $1/\beta_i$ increases. Similarly, this probability increases as $r_j$ increases and it approaches $0.5$ as $r_j$ decreases. This means that $1/\beta_i$ indicates the difficulty level of problem $x_i$, and $r_j$ shows the expertise level of user $u_j$ ($r_j=0$ indicates that $u_j$ is a spammer and $r_j<0$ indicates that $u_j$ is an adversary). 

Moreover, Gaussian prior distribution is considered over $r_j$. In addition, 
in order to ensure that $\beta_i$ can only take positive values, it was re-parametrized as $e^{\beta'}$ and a Gaussian distribution is considered as the prior over $\beta'$. GLAD's ability to find spammers and adversary users, accompanied by its capability to model users with different levels of ability and problems with different levels of difficulty, makes it a suitable choice for label aggregation in real datasets. Moreover, its data complexity is low. However, it can not distinguish user expertises in labeling different categories of problems. An equivalent method to GLAD is also proposed in \cite{AIAAI}, which assumes that $\beta_i$s are known.

DARE is another method for modeling both user expertises and problem difficulty levels \cite{DARE}. The method assumes that $\boldsymbol{\theta_j}=r_j$ and $\boldsymbol{\beta_i}=\{\beta_i, \delta_i\}$, where $r_j$ and $\beta_i$ respectively indicate the expertise of user $u_j$ and difficulty level of problem $x_i$, and $\delta_j$ determines how discriminative $x_i$ is. 
DARE assumes that if the user $u_j$ doesn't know the true answer of a question, she will answer it randomly. Moreover, according to this model, the user knows the true label of $x_i$ with probability $p_{ij}=\Phi(\sqrt{\delta_i}(r_j-\beta_i))$ where $\Phi$ is the Gaussian cumulative distribution function, i.e., 
\begin{equation}
P(A_{ij}=y_i|r_j, \beta_i, \delta_i)=p_{ij}+\frac{1}{C}(1-p_{ij})
\end{equation}
where $C$ is the number of classes. In order to make the inference tractable, Gaussian priors are assumed over both user expertise and problem difficulty and Gamma distribution was chosen as the prior over discriminations. 
DARE is a complex model and is suitable for the case that problems are very diverse in difficulty and ambiguity levels, and also the users answer to the best of their abilities, and hence, significant results are not expected in most general crowd labeling applications when using this method.

In \cite{LFM}, the GLAD method is extended. This method assumes that many other factors may be important in label aggregation. In this method, the difficulty of problem $x_i$ and expertise of user $u_j$ are respectively modeled by a set of random variables $\boldsymbol{\beta_i}$ and $\boldsymbol{\alpha_j}$.  It is also assumed that these factors are not pre-defined, but they emerge naturally from the data. In addition, a variable $\gamma_j$ is also considered for user $u_j$, which shows her bias. According to these assumptions, the label generation model is revised to:
\begin{equation}
P(A_{ij}=y_i|y_i, \boldsymbol{\alpha_j}, \gamma_j, \boldsymbol{\beta_i}) = \frac{1}{1+e^{\boldsymbol{\alpha_j}^T. \boldsymbol{\beta_i}+y_i\gamma_j}}
\end{equation}
Although these factors are able to capture a wide range of variability in the correctness probabilities of labels, but since they are not pre-defined, they do not necessarily have high-level semantic meaning. Moreover, these latent factors don't share attributes across problems or labelers and lots of user labels are required to estimate their values. In order to solve these issues, the method assumes that there are sets of pre-defined features to be used in modeling problems and labelers. For example, demographic properties such as age, sex, and level of education are samples of such predefined features for labelers. In addition, the method assumes that the value of the features are known for all problems and lebelers. Using the provided features, the method models the latent factor vectors as a linear combination of the pre-defined features and an unknown set of weights. That is, $\boldsymbol{\alpha_j} = \boldsymbol{\Phi_{\boldsymbol{\alpha_j}}}\boldsymbol{W_\alpha}$, $\boldsymbol{\beta_i} = \boldsymbol{\Phi_{\boldsymbol{\beta_i}}}\boldsymbol{W_\beta}$, and $\gamma_j = \boldsymbol{\Phi_{\gamma_j}}\boldsymbol{w_\gamma}$, where $\boldsymbol{\Phi}$ denotes the values of features set and $\boldsymbol{W}$ ($\boldsymbol{w}$) is the weight matrix (vector) relating the latent factors to the features set. The main issue of this approach is the need to define the features sets and also providing their values for all labelers and problems. An almost identical method is proposed in \cite{MDimWCrowd}.

\section{Inference algorithms}
As it was disscussed in Section \ref{Sec:LatentModels}, crowd labeling models can be formulated under a unified probabilistic framework. In this framework, each model corresponds to a generative process of the data using a set of latent random variables. This process specifies the joint probability distribution of hidden and observed random variables. Random variables can be categorized into three groups. Observed variables ($x$), e.g. provided labels $\boldsymbol{A}$, query latent variables whose values we wish to know ($y$), e.g. the true labels of samples $\boldsymbol{Y}$, and the nuisance latent variables ($z$), e.g. user capabilities $\boldsymbol{\theta}$ \cite{Seeger-BayesReport}. The common goal of all disscussed methods is inferring the hidden structure that most likely generated the observed data. There are different approaches for making inference on these models and estimating the values of desirable latent variables. 

In a fully Bayesian approach, the posterior distribution of the desired random variables given observed variables is computed by marginalizing over the nuisance random variables and conditioning on the observed variables, i.e.
\begin{eqnarray}
p(y|x)=\frac{\sum_{z}^{}{p(x,y,z)}}{p(x)}
\end{eqnarray}
Here we assumed that the latent variables are discrete. In the continuous case, the summations are replaced by integration. Although this approach is quite simple since it only uses marginalization and conditioning, the required integrals are analytically intractable in most models, such as crowd categorization ones. However, utilizing independence relations among random variables can facilitate this process. Graphical models provide a graphical representation for the dependency relations between random variables. Belief Propagation (BP) \cite{BelProp} is a Bayesian inference algorithm on graphical models which utilizes the independence relations to reduce computational complexity. BP is a message-passing algorithm. It operates by sending local messages (the beliefs) between adjacent variables in the graphical model. The message $v_{i\rightarrow j}$ is the belief of $v_i$ of what $v_j$ should be, which is based on messages to $v_i$ of all neighbors except $v_j$. The messages are updated iteratively and in a parallel manner. A BP-like algorithm for crowd categorization is heuristically introduced in \cite{SVD1} for the model depicted in Fig. \ref{DSLM}. Let $x_{i\rightarrow j}$ and $u_{j\rightarrow i}$ be real-valued messages from questions to labelers and from labelers to questions, respectively. Initializing $u_{j\rightarrow i}^0$ randomly from Gaussian distribution $\mathcal{N}(1, 1)$ or deterministically by $u_{j\rightarrow i} = 1$, the algorithm updates the messages at $t$-th iteration via, 
\begin{equation}
\label{BPlike}
x_{i\rightarrow j}^{t+1}=\sum_{j' \in \boldsymbol{\partial_i}\backslash j}{A_{ij'}u_{j' \rightarrow i}^t}, \qquad u_{j\rightarrow i}^{t+1}=\sum_{i' \in \boldsymbol{\partial_j}\backslash i}{A_{i'j}x_{i' \rightarrow j}^{t+1}}
\end{equation}
and the labels are estimated via $\hat{s}_i^t=sign\left[\hat{x}_i^t\right]$, where $\hat{x}_i^t=\sum_{j \in \boldsymbol{\partial_i}}{A_{ij}u_{j \rightarrow i}^t}$, $\boldsymbol{\partial_l}$ denotes all neighbours of $l$ in the graph, and $\boldsymbol{S}\backslash k$ excludes $k$ from the set $\boldsymbol{S}$. $x$ and $y$ play the roles of weighted majority votes and user reliabilities, respectively. In \cite{VarInf}, BP is used to make inference on the graphical model depicted in Fig. \ref{DSLM}. The authors show that the rule set in (\ref{BPlike}) is a special case of their algorithm when Haldane prior distribution is assumed as the prior over user reliabilities.

Singular Value Decomposition (SVD) is also beneficial to draw inferences from crowd categorization of binary problems. When $\boldsymbol{A}$ is a $(l, r)$-regular bipartite graph with $l=r$, the rules in (\ref{BPlike}) are very similar to the ones in power iteration \cite{PowerIteration} which is a method to compute the leading singular vectors of a matrix. At the presence of matrix $\boldsymbol{A}_{N \times R}$ and two vectors $\boldsymbol{u} \in \mathbb{R}^N$ and $\boldsymbol{v} \in \mathbb{R}^R$, power iteration starts with a random initialized $\boldsymbol{v}$, and then it iteratively updates $\boldsymbol{u}$ and $\boldsymbol{v}$ according to:
\begin{equation}
\forall i, \; u_i=\sum_j{A_{ij}v_j}, \qquad \forall j, \; v_j=\sum_i{A_{ij}u_i}
\end{equation}
It is known that randomized $\boldsymbol{u}$ and $\boldsymbol{v}$ converge linearly to the leading left and right singular vectors. 
Inspired by this similarity, the following algorithm is proposed for binary categorization problems in \cite{SVD2}:
\begin{enumerate}
\item Compute the left and right singular vector of $\boldsymbol{A}$, corresponding to the top singular values of $\boldsymbol{A}$.
\item Since both $(\boldsymbol{u}, \boldsymbol{v})$ and $(-\boldsymbol{u}, -\boldsymbol{v})$ are valid pairs of leading singular vectors, the mass of the element values is considered to be able to resolve the ambiguity, if $\sum_{j:v_j \geq 0}{v_j^2} \geq \sum_{j:v_j < 0}{v_j^2}$, then $\hat{y}_i=sign(u_i)$, otherwise $\hat{y}_i=sign(-u_i)$. 
\end{enumerate}
Note that the algorithm and the power iteration rules are not exactly the same. In the updating rules attributed to the algorithm, the received signals from the destination will be excluded (`$\backslash j$'s in the algorithm). But, these signals are presented in the power iteration. In other words, the power iteration rules are indeed the simplified versions of those in the algorithm, because the latter approximates all different $u_{i \rightarrow j}$ with a common $u_i$. 

Although BP reduces the computational complexity of Bayesian inference, still making inference in more complex models such as Fig. \ref{DARELM} is intractable. Hence, approximate inference algorithms are the favored solutions in these cases. Approximate Bayesian inference algorithms can be categorized into deterministic and nondeterministic classes. Nondeterministic methods are based on sampling methods such as \textit{Markov Chain Monte Carlo} (MCMC) \cite{MCMC}. Although these methods are broadly applicable and can be applied to a wide range of distributions, in practice, Monte Carlo methods are computationally expensive and their computational demands often limit their use to small scale problems. Therefore, using sampling based methods such as MCMC are not very common in crowd categorization problems in which the size of the datasets are usually large and hence, deterministic approaches such as variational methods have received great attention in this area. Variational approximation is a family of inference algorithms that approximate the desired posterior $p(y|x)$ with a member of a family of distributions $q(y)$, working with which is tractable, e.g. Gaussian distributions \cite{BishopPRML}. In these methods, the measure for selecting the most appropriate $q(y)$ from the set of candidate distributions is a member of the \textit{alpha family} of divergences \cite{alphafamily}. Mean field approximation, is one of the most applicable variational methods \cite{BishopPRML}. It approximates the posterior density $p(y|x)$ by a factorized density function $q(y)=\prod_i{q_i(y_i)}$ and iteratively optimizes each of the base density functions such that their product becomes closer to the desired posterior. Mean field measures the distance between the approximated function and the posterior using
\begin{eqnarray}
KL(q||p)=-\sum_{y}{q(y) \ln \left\lbrace \frac{p(y|x)}{q(y)} \right\rbrace}
\end{eqnarray}
Kullback-Leibler divergence measures the distance between the approximated function and the posterior. Another variational inference framework is Expectation Propagation (EP) \cite{MinkaUAI,MinkaPhD}. Similar to mean field approximation method, this method is also based on minimization of KL-divergence between approximated function and the posterior but of the reverse form (Note that the KL divergence is not symmetric). This makes the approximation rather different. Mean field is zero-forcing, i.e. underestimates the posterior variance, whereas EP is zero-avoiding, i.e. the approximated function overestimates the posterior variance \cite{BishopPRML}.
In \cite{VarInf}, mean field approximation is used to make inference on the graphical model presented in Fig. \ref{DSLM}. This method approximates the posterior distribution $P(\boldsymbol{Y},\boldsymbol{\alpha}|\boldsymbol{A})$ by the factorized distribution $q(\boldsymbol{Y},\boldsymbol{\alpha})=\prod_{i=1}^N {f_i(y_i)\prod_{j=1}^R{g_j(\alpha_j)}}$, where $\alpha_j$ is the accuracy of user $u_j$. This method uses block coordinate descent method to estimate factors $f_i(y_i)$s and $g_j(\theta_i)$s that alternatively optimizes the factors. The authors show that their approximation method is closely related to EM algorithm, which will be described in the sequel. Moreover, in \cite{DARE} an approximation inference method based on EP is used to calculate the marginal distributions on the underlying factor graph corresponding to the model depicted in Fig. \ref{DARELM} by iteratively calculating messages along edges that propagate information across the factor graph.

Another approach for estimating the latent variables of probabilistic models is finding a point estimate for query latent variables by optimizing an objective function such as the log-likelihood function,
\begin{eqnarray}
\hat{Y}=arg \max_{y}{\log{P(x|y)}}
\end{eqnarray}
In order to find likelihood function, we have to marginalize over nuisance variables. Since this is time consuming, a common method Expectation Maximization (EM) is used. EM is a two stage iterative algorithm for finding maximum likelihood estimates. In crowd categorization models, EM iteratively (i) estimates the true labels according to the current estimates of model parameters, and (ii) re-estimates the model parameters based on the current beliefs about the true labels. For example, in \cite{EM1979} the likelihood is, 
\begin{equation} 
L=P(\boldsymbol{A}|\boldsymbol{\Pi}, \boldsymbol{P})= \prod_{i=1}^N{\left( \sum_{k=1}^J{P(C_k) \prod_{j=1}^R{\prod_{l=1}^J{\left( \pi_{kl}^j \right)^{n_{il}^j}}}} \right)} 
\end{equation} 
where $\boldsymbol{\Pi}$ is the set of all user's confusion matrices ($\pi_{ik}^j$s are the elements of the confusion matrix of users $u_j$), $\boldsymbol{P}=\{P(C_l)\}_{l1}^J$ is the set of all class prior probabilities, and $n_{il}^j$ is the number of labels $C_l$ which is assigned to problem $x_i$ by users $u_j$. Also, $T_{iq}=1$, if the label $C_q$ is a true one for the problem $x_i$, and it is zero, otherwise. Maximization of $L$ is a complicated task and hence, EM algorithm helps to estimate the parameters. In E step, true labels are estimated by user's confusion matrices and prior probabilities included in the previous step. $C_k$ is the label of $x_i$ with the following probability:
\begin{eqnarray}
\label{EM792}
P(T_{ik}=1|\boldsymbol{\Pi}, \boldsymbol{P})&\propto& \prod_{j=1}^R{\prod_{l=1}^J{\left( \pi_{kl}^j\right)^{n_{il}^j} P(C_k)}}
\end{eqnarray}
Maximizing the likelihood through the estimated labels in M step leads to the following estimations for parameters: 
\begin{equation}
\label{EM791}
\hat{\pi}_{kl}^j=\frac{\sum_{i=1}^N{T_{ik}n_{il}^j}}{\sum_{l=1}^J{\sum_{i=1}^N{T_{ik} n_{il}^j}}}, \qquad \hat{P}(C_k)=\frac{\sum_{i=1}^N{T_{ik}}}{N}
\end{equation}
Based on prior probabilities for parameters in the latent model, the joint probability can replace the likelihood in EM. 

As the last category of methods, we describe methods which use Matrix Factorization (MF) to estimate missing (unseen) values of sparse matrices before making inferences. The goal of MF is to factorize the matrix $\boldsymbol{A}$, for example as factors $\boldsymbol{U}_{k \times N}$ and $\boldsymbol{V}_{k \times R}$, where
\begin{equation}
\label{MF}
\{\boldsymbol{U},\boldsymbol{V}\} = arg\min_{\boldsymbol{X},\boldsymbol{Y}} \|\mathcal{P}_\Omega(\boldsymbol{A}-\boldsymbol{X}^T\boldsymbol{Y})\|_F^2
\end{equation}
and $\mathcal{P}_\Omega(\cdot)$ indicates the observed elements of the given matrix. Then, $\boldsymbol{A'}=\boldsymbol{U}^T\boldsymbol{V}$ contains the observations and estimates of unseen entries. The described SVD-based method in the earlier paragraphs is based on MF. It approximates the target matrix with a rank-$1$ matrix, but it uses sum-squared distance between all entries of target and estimated matrices. When $\boldsymbol{A}$ is sparse, the distance should only be computed for the observed entries of the target matrix. This may lead to a complex non-convex optimization problem \cite{WLRA}. In such situation, it is possible to use Probabilistic Matrix Factorization (PMF) which is a well-known approach in collaborative filtering \cite{PMF}. It induces a latent feature vector for each person and example, in order to infer unobserved user ratings for all examples. In a similar way, PMF is used in crowd categorization to estimate the unseen labels. PMF can be considered as an inference algorithm on the model depicted in Fig. \ref{DARELM}. In PMF, column vectors $\boldsymbol{U_i}$ and $\boldsymbol{V_j}$ respectively represent $k$-dimensional labeler-specific and question-specific latent feature vectors, and the conditional distribution over the collected labels, using the Gaussian distribution, is defined as, 
\begin{eqnarray}
P(\boldsymbol{A}|\boldsymbol{U}, \boldsymbol{V}, \sigma^2) = \prod_{i=1}^N{\prod_{j=1}^R{\mathcal{N}(A_{ij}|\boldsymbol{U_i}^T\boldsymbol{V_j}, \sigma^2)^{I(A_{ij}\neq 0)}}}
\end{eqnarray}
Also, zero-mean spherical Gaussian priors are considered for latent feature matrices, i.e. 
\begin{eqnarray}
P(\boldsymbol{U}|\sigma_U^2)=\prod_{i=1}^N{\mathcal{N}(\boldsymbol{U_i}|0, \sigma_U^2\boldsymbol{I})}, \quad P(\boldsymbol{V}|\sigma_V^2)=\prod_{j=1}^R{\mathcal{N}(\boldsymbol{V_j}|0, \sigma_V^2\boldsymbol{I})}
\end{eqnarray}
To estimate model parameters, PMF maximizes the log-posterior distribution over feature matrices with hyper-parameters. The maximization is equivalent to minimizing the following squared error with $L_2$ regularization, 
\begin{equation}
\label{PMFEq}
\frac{1}{2}\sum_{i=1}^N{\sum_{j=1}^R{I(A_{ij} \neq 0)(A_{ij}-\boldsymbol{U_i}^T\boldsymbol{V_j})^2}}+\frac{\lambda_U}{2}\|\boldsymbol{U}\|_F^2+\frac{\lambda_V}{2}\|\boldsymbol{V}\|_F^2
\end{equation}
where $\lambda_U=\frac{\sigma_U}{\sigma}$, $\lambda_V=\frac{\sigma_V}{\sigma}$, and $\|\centerdot\|_F^2$ denote the Frobenius norm. $\boldsymbol{U}$ and $\boldsymbol{V}$ are found by gradient descent algorithm, and missing values of $\boldsymbol{A}$ are inferred by taking the inner product of estimated $\boldsymbol{U}$ and $\boldsymbol{V}$. The equation in (\ref{PMFEq}) is the same equation as in (\ref{MF}) with an additional regularization term. Finally, It should be noted that any inference algorithm can be used to estimate the true labels from the collected and estimated unseen labels. For example, majority voting as well as \cite{EM1979} methods are used in \cite{PMFC1,PMFC2}, respectively. 

In Table \ref{summary} (in conclusion section), you can see the inference algorithm types of all major surveyed method.

\section{Cost efficient methods}
Crowd labeling methods can be categorized into three main groups: one-shot, inductive, and adaptive. So far, all methods that have been discussed are one-shot. That is, a number of labels are collected for a set of problems and the true labels are estimated by aggregating the user labels. 

Inductive methods, collect a set of labels for a limited number of samples and use the aggregated labels to train a classifier that can predict the label of any other sample \cite{EvilTeacher,VoxPopuli,LFCE,ConvLFC,LiesABit,CrowLearn,KnowsABit}. Moreover, these methods can also collect labels in an active manner. That is, each sample is either classified by the classifier, or its label is estimated using crowd labeling and is used to improve the classifier \cite{GALabel,Vuvuzela,STALC}. In inductive methods, each sample must be described as a feature vector. Moreover, a criterion is required to decide whether to acquire the label from the classifier or the crowd \cite{SSAL}. Hence, the success rate of inductive methods depends on: (i) the transformation method from the object space to the feature space, (ii) the type of classifier model, (iii) the decision criterion, and (iv) the quality of estimated labels using collected labels from the crowd. The first three factors are out of the scope of this paper. Therefore, we do not survey inductive methods.

Adaptive methods spend budget economically and collect the labels iteratively. These methods collect a set of labels in each step, based on the assumed latent model and current collected labels. They use the new acquired labels to re-estimate the true labels and update the latent model. That is, the updated model in each step participates in selecting samples in the next step. It is obvious that an oracle procedure selects samples for which acquiring new labels maximizes the overall performance. In the rest of this section, we describe adaptive methods for crowd labeling.

In this section, we assume that the labeling cost is equal for all samples, and only one label is requested for one of the samples in each step, based on a sample selection criterion.
The simplest criterion selects the sample with minimum number of currently collected labels. The result is an approximately equal number of labels for all samples. This criterion is called ``uniform sample selection". As samples with low difficulty levels need fewer labels, in most cases uniform criterion wastes the budget. 

Heterogeneity of collected labels is a non-uniform criterion. It selects a sample that enjoys the minimum heterogeneity of its current collected labels. 
Considering the same level of reliabilities for all users and uniform prior over possible labels, heterogeneity of collected labels for sample $x_i$ can be measured using the entropy of $P(y_i|\boldsymbol{A_i})$, which follows a beta distribution \cite{GALabel,NoisyLabelers}. Therefore, heterogeneity criterion selects sample $x_{i^*}$ to request a new label where,
\begin{equation}
\label{Ent}
i^*=arg \max_{i}\; H(y_i|\boldsymbol{A}_i) =arg \max_{i}\;\left\lbrace -\sum_y{P(y_i=y|\boldsymbol{A}_i)\log P(y_i=y|\boldsymbol{A}_i)} \right\rbrace
\end{equation}
Entropy has a bias toward selecting the samples with more labels. For example, in first steps it never selects samples with only one label in their current labels set. Therefore, using entropy results in having a few samples with many labels, and several samples with few labels. Moreover, since entropy doesn't consider the labeler capabilities, it doesn't request more labels for samples with heterogeneous wrong labels.

The uncertainty in estimated labels is another non-uniform criterion for sample selection. In \cite{GALabel} uncertainty is defined as the probability that the majority label of a sample is not true, i.e.,
\begin{equation}
\label{Unc}
i^*=arg \max_{i}\;\left\lbrace 1- P(y_i=\hat{y_i}) \right\rbrace
\end{equation}
where $\hat{y_i}=arg \max_{y} \left\lbrace P(y_i=y|\boldsymbol{A}) \right\rbrace$, if it is assumed that all labelers have the same reliabilities and the considered prior distribution over possible labels is uniform. In this configuration, $P(y_i|\boldsymbol{A_i})$ follows a Beta distribution $\beta(a+1,b-1)$ where $a$ and $b$ are the number of collected labels from different categories. Therefore,
\begin{eqnarray}
P(y_i=\hat{y}_i)=\max\{I_{0.5}(a+1,b+1), I_{0.5}(b+1,a+1)\}
\end{eqnarray}
where $I_x(m,n)$ is the cumulative distribution function of the Beta distribution.

Both entropy and uncertainty criteria can be extended to consider user reliabilities by using more complex models described in Section \ref{Sec:LatentModels}. The main weakness of both of these criteria is that they choose samples only based on current collected labels, i.e., they are blind to the result of their decisions. The methods that are aware of their decisions must measure the effect of selecting a sample by using a defined indicator. The proposed criterion in \cite{DARE} defines the uncertainty in estimates of user model parameters as the indicator. It selects a sample which acquiring a new label is expected to maximize the reduction in the defined indicator. The method measures the uncertainty in estimates of model parameters by the entropy of posterior distributions of the parameters. The method assumes that $P(\alpha_j|\boldsymbol{A})= \mathcal{N}(\mu_j,\sigma_j)$ and $P(\alpha_j|\boldsymbol{A},a_{ij})= \mathcal{N}\left(\mu_j(a_{ij}),\sigma_j(a_{ij})\right)$, where $\alpha_j$ is the accuracy of user $u_j$, and $a_{ij}$ is the next provided label by the crowd. Since $a_{ij}$ is not known, the criterion uses the expectation, i.e. the criterion selects sample $x_{i^*}$ to request a new label where, 
\begin{equation}
i^*=arg \max_{i}\; \left\lbrace \mathbb{E}_{a_{ij}|\boldsymbol{A}}\; \left\lbrace H(\alpha_j|\boldsymbol{A}) -H(\alpha_j|\boldsymbol{A},a_{ij}) \right\rbrace \right\rbrace
\end{equation}
where $H$ indicates the entropy function. It can be shown that if Gaussian distributions are considered over user model parameters, then \cite{DARE},
\begin{equation}
H(\alpha_j|\boldsymbol{A})-H(\alpha_j|\boldsymbol{A},a_{ij}) = \frac{1}{2} \ln \frac{\sigma_j^2}{\sigma_j^2(a_{ij})}
\end{equation}
This criterion, which is called as \textit{AccsExpEntRed} in the experiments, leads to find more accurate user model parameters.

To focus on more accurately estimated labels, the future indicator can be defined as the uncertainty in estimates of true labels. The uncertainty can be measured by using the incorrectness probability of the estimated labels. The criterion selects sample $x_{i^*}$ to request a new label where,
\begin{equation}
i^*=arg \max_{i}\; \left\lbrace\mathbb{E}_{a_{ij}|\boldsymbol{A}}\;\left\lbrace \max_y P(y_i=y|\boldsymbol{A},a_{ij})-\max_y P(y_i=y|\boldsymbol{A}) \right\rbrace \right\rbrace
\end{equation}

The main goal of sample selection is to identify challenging problems and spend money efficiently when collecting labels. A criterion to distinguish difficult problems from simple ones is that getting a new label for challenging samples doesn't necessarily reduce the uncertainty in their estimated labels. Specially, when there are many unreliable labelers and the observations are not sufficient to accurately estimate their capabilities. Therefore, a method to recognize difficult problems is to choose a sample for which getting a new label is expected to maximize the uncertainty in its estimated label, i.e., such a criterion selects sample $x_{i^*}$ to request a new label where,
\begin{equation}
i^*=arg \max_{i}\;\left\lbrace \mathbb{E}_{a_{ij}|\boldsymbol{A}}\;\left\lbrace 1-\max_y P(y_i=y|\boldsymbol{A},a_{ij}) \right\rbrace \right\rbrace
\end{equation}
We refer to the two proposed criteria as \textit{ExpAccUnc} and \textit{AccUnc} in the experiments.

\section{Experimental results}
\subsection*{Test setup}
We implemented the major surveyed methods, then we compared them using human and synthetic datasets. All reported results are averaged over 10 runs. Available labels for each samples are shuffled in each run, and all methods are used the same shuffled data in that run. Every time an algorithm requests a label for a sample, the first unseen label of that sample is returned. After getting on the average one new label per each sample, the accuracy of the method is re-calculated. In addition, the labels are converted to ordered ranks, in Probabilistic Matrix Factorization (PMF). Each data matrix is factorized by using decomposition ranks $k=3,5,10,20,30,50$ and regularization factors $\lambda = 0.001,0.01,0.1,0.3,0,5$. Afterwards, the factors with lowest estimation error are selected as the result. The estimation error is considered as the mean squared errors of the known values and their corresponding estimates. Moreover, in the PMF's iterative algorithm $U$ and $V$ are initialized with small Gaussian random values. In addition, the algorithm stops when error becomes less than $0.01$ or number of iterations reaches $1000$. Furthermore, the number of iterations in Belief Propagation (BP) is considered as the average number of labels per sample. Moreover, BP stops when the estimated labels remain unchanged from the previous step, or the number of iterations reaches $100$. 

The implemented methods are listed in Table \ref{ImplementedMethods}. The abbreviation column indicates the short names given to methods which appear in subsequent plots and result tables. Note that since models with multidimensional parameters \cite{MDimWCrowd,LFM} require external information about labelers' or problems' features, we did not consider them in our experimental comparisons. 
\begin{table}
\centering
\begin{tabular}{cc}
\textbf{Title} & \textbf{Citation} \\
\cline{1-2} MV & Majority voting \\
\cline{1-2} Accuracy & \cite{txteagle} \\
\cline{1-2} ConfMatrix & \cite{EM1979} \\
\cline{1-2} DARE & \cite{DARE}\\
\cline{1-2} BP & \cite{SVD1} \\
\cline{1-2} GLAD & \cite{GLADPaper} \\
\cline{1-2} PMF MV & \cite{PMFC1}\\
\cline{1-2} PMF ConfMatrix & \cite{PMFC2} \\
\cline{1-2} Entropy & \cite{GALabel}\\
\cline{1-2} Uncertainty & \cite{GALabel}\\
\cline{1-2} AccsExpEntRed & \cite{DARE}\\
\cline{1-2} AccUnc & Proposed by the authors\\
\cline{1-2} ExpAccUnc & Proposed by the authors\\
\cline{1-2} 
\end{tabular} 
\caption{Implemented methods used in experimental comparisons. The first 8 methods are one-shot, while the other 5 ones are adaptive. BP, GLAD and PMF-based methods only work with binary problems, while other methods handle all cases.}
\label{ImplementedMethods}
\end{table}

\subsection*{Datasets}
\begin{figure}
\centering
\subfigure[RTE]{
\includegraphics[scale=0.7]{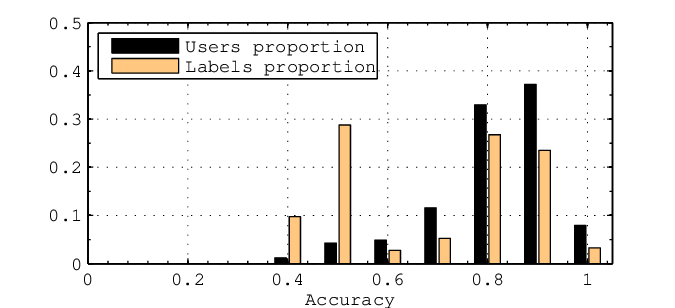}
}
~
\subfigure[TEMP]{
\includegraphics[scale=0.7]{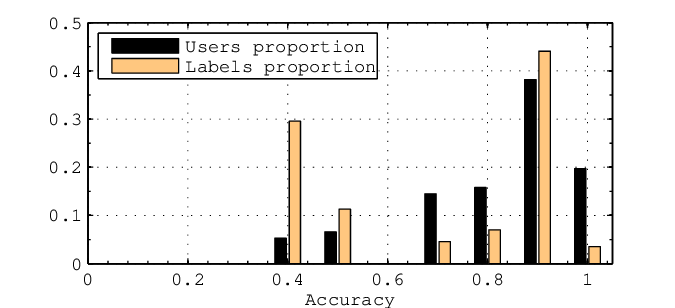}
}
~
\subfigure[Duchenne]{
\includegraphics[scale=0.7]{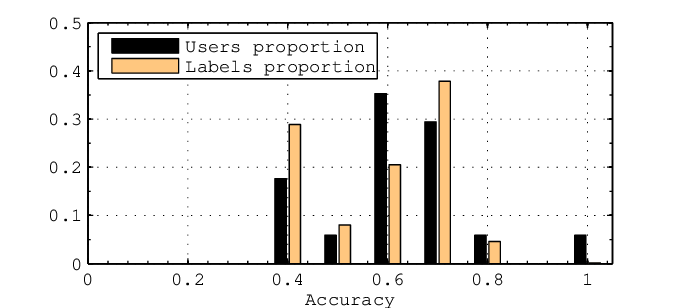}
}
\caption{The histograms of user and label accuracies for human datasets.}
\label{DSHists}
\end{figure}

The implemented methods are compared using the following datasets:
\begin{itemize}
\item \textbf{Recognizing Textual Entailment (RTE).} In RTE dataset \cite{CheapFast}, users are given two sentences and a binary choice of whether the second hypothesis sentence can be inferred from the first one, or not. The human responses are provided by the MTurk users. 
\item \textbf{Temporal event recognition (TEMP).} In TEMP dataset \cite{CheapFast}, the users must choose one of the two labels ``strictly before" or ``strictly after" to represent the temporal relation between two event-pairs. MTurk users labelled the problems of the dataset.
\item \textbf{Duchenne.} In each face image of the Duchenne dataset \cite{GLADPaper}, MTurk users are asked to determine whether the face contains Duchenne smile (``enjoyment" smile) or not (a ``social" smile). Before using the dataset we eliminated samples without crowd labels or ground truth. Moreover, we removed repetitive or incompatible crowd responses.
\item \textbf{Synthetic datasets.} Multiple synthetic datasets are generated to evaluate methods on multi-class problems. Synthetic datasets are formed based on the above binary human datasets. The synthetic datasets are so randomly generated that have the same number of problems and workers, and the same user accuracies' and label accuracies' histograms as the base ones. We generated 3-class, 4-class and 5-class versions of each dataset. 
\end{itemize}
Table \ref{DSsInfo} contains the properties of the human datasets. Fig. \ref{DSHists} depicts the ratio of labelers with various levels of accuracies, for each dataset. The ratio of provided labels by the labelers with various levels of accuracies are also depicted in the figure. These  properties indicate that the Duchenne is the most challenging dataset, because there are fewer crowd labels per sample; it is full-rank; adversary and unreliable labelers have provided most of the collected labels; and the ratio of correct crowd labels to all crowd labels is less than other datasets. In a similar manner, it may be inferred that TEMP is the easiest dataset.
\begin{table}
\centering
\begin{tabular}{cccc}
& \textbf{RTE} & \textbf{TEMP} & \textbf{Duchenne} \\
\cline{1-4} 
\textbf{Number of Samples} & 800 & 462 & 159 \\
\cline{1-4} 
\textbf{Number of Labelers} & 164 & 76 & 17 \\
\cline{1-4} 
\textbf{Labels per Sample} &10 & 10 & 4-10\\
\cline{1-4} 
\textbf{Total Labels} & 8000 & 4620 & 1150 \\
\cline{1-4} 
\textbf{Ratio of Correct Labels} & 0.729 & 0.734 & 0.624 \\
\cline{1-4} 
\textbf{Rank of data Matrix} &148 & 61 & 17\\
\cline{1-4} 
\end{tabular} 
\caption{The properties of the human datasets.}
\label{DSsInfo}
\end{table}

\subsection*{Results and Analyses}
\begin{figure}
\centering
\subfigure[2classes-RTE]{
\includegraphics[scale=0.30]{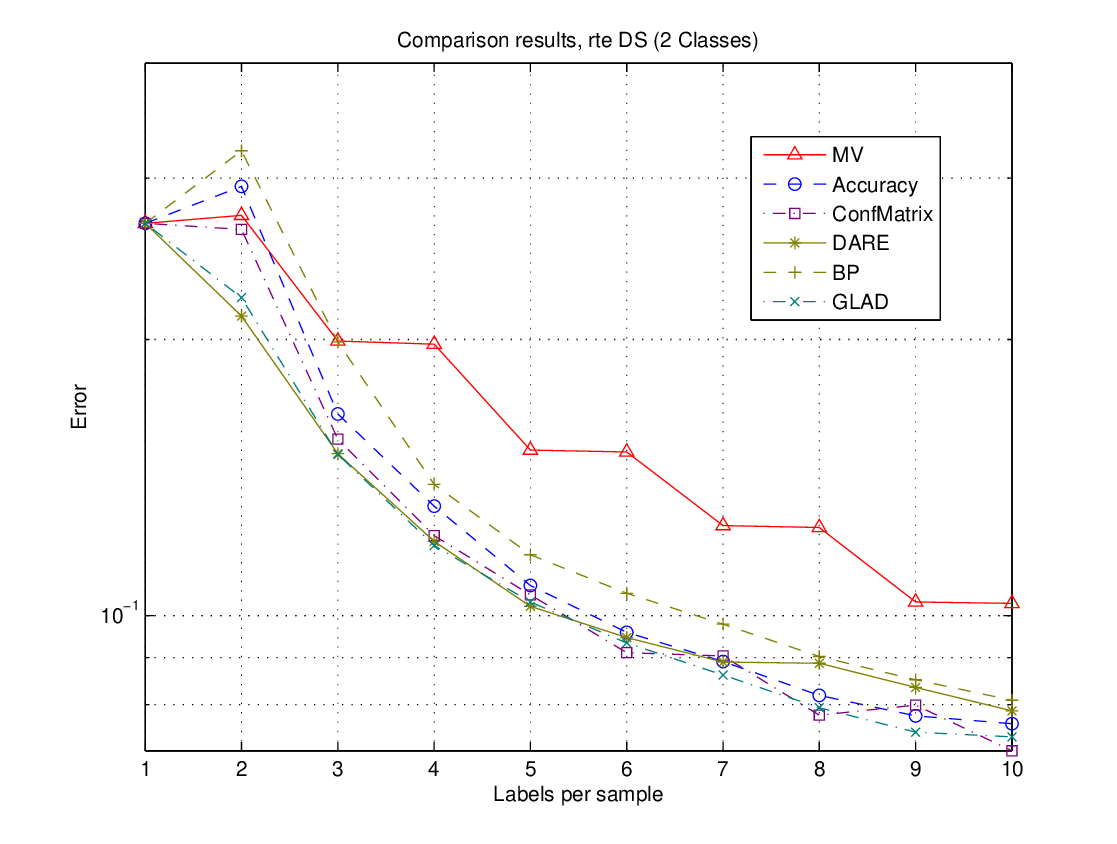}
}
~
\subfigure[3class-RTE]{
\includegraphics[scale=0.40]{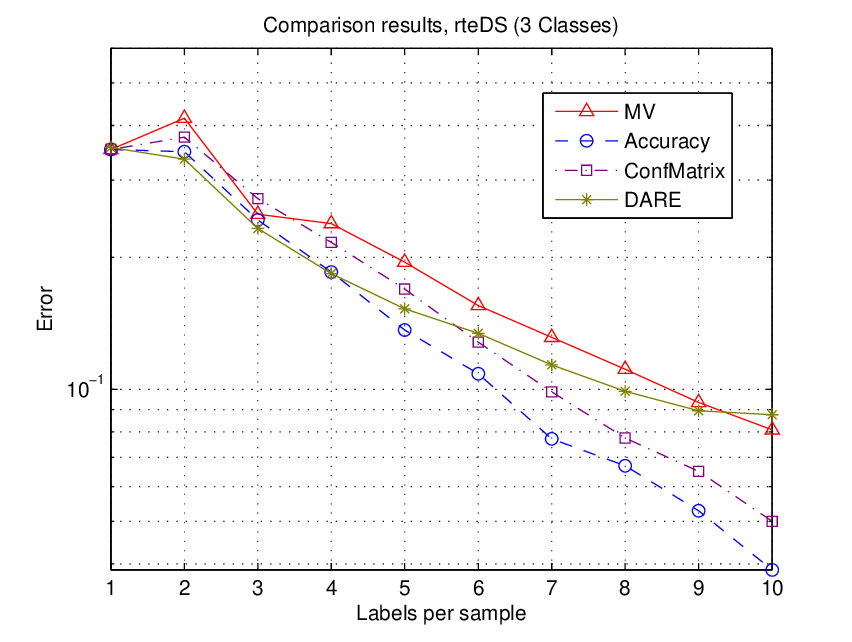}
}
\caption{Results of one-shot methods on 2 and 3-class samples (RTE dataset).}
\label{AllMethods_rte}
\end{figure}

The results of comparing different methods on three datasets 2-class-Duchenne, 3-class-RTE and 4-class-TEMP datasets as samples are shown in Tables \ref{DuResults}, \ref{rteResults3} and \ref{TempResults4}, respectively\footnote{To access the source codes, datasets, and all other results, please refer to the first author's web page at http://ce.sharif.edu/$\sim$muhammadi.}. In all tables, the best results are given in bold. In addition, the results of one-shot methods on 2 and 3-class samples for RTE dataset are depicted in Fig. \ref{AllMethods_rte}. 

\begin{table}
\centering
\begin{tabular}{ccccc}
\textbf{Labels/Sample} & \textbf{1} & \textbf{2} & \textbf{3} & \textbf{4}  \\
\cline{1-5} MV &   			62.08  &  61.98  &  66.54  &  66.42 \\
\cline{1-5} Accuracy &   	62.08  &  61.01  &  69.87  &  \textbf{72.58} \\
\cline{1-5} ConfMatrix &   	62.08  &  63.96  &  69.62  &  71.51 \\
\cline{1-5} DARE &   		62.08  &  \textbf{66.48}  &  \textbf{71.38}  &  \textbf{73.52} \\
\cline{1-5} BP &   			62.08  &  59.62  &  68.43  &  72.14 \\
\cline{1-5} GLAD &   		62.08  &  \textbf{65.28}  &  \textbf{70.13}  &  71.82 \\
\cline{1-5} PMF MV &   		46.54  &  50.63  &  57.99  &  62.01 \\
\cline{1-5} PMF ConfMatrix &45.97  &  52.20  &  57.92  &  63.84 \\
\cline{1-5} Entropy &   	62.08  &  \textbf{64.59}  &  68.96  &  70.88 \\
\cline{1-5} Uncertainty &   62.08  &  62.11  &  68.02  &  70.60 \\
\cline{1-5} AccsExpEntRed & 59.62  &  \textbf{66.10}  &  \textbf{70.44}  &  \textbf{73.33} \\
\cline{1-5} AccUnc &   		62.08  &  \textbf{65.09}  &  \textbf{71.19}  &  \textbf{73.77} \\
\cline{1-5} ExpAccUnc &  	59.50  &  \textbf{65.09}  &  \textbf{70.13}  &  \textbf{72.39} \\
\cline{1-5} 
\end{tabular} 
\caption{Results of 2-class-Duchenne dataset. Rows indicate implemented methods and columns indicate the average number of labels that are used for each problem.}
\label{DuResults}
\end{table}
\begin{table}
\centering
\begin{tabular}{ccccccccccc}
\textbf{Labels/Sample} & \textbf{1} & \textbf{2} & \textbf{3} & \textbf{4} & \textbf{5} & \textbf{6} & \textbf{7} & \textbf{8} & \textbf{9} & \textbf{10} \\ 
\cline{1-11}MV   			& 64.71   & 58.43   & 74.89   & 76.11  &  80.48  & 84.47  &  86.85  &  88.89  &  90.66  &  91.92\\
\cline{1-11}Accuracy   		& 64.71   & 65.14   & 75.64   & 81.48  &  86.34  &  89.14  &  \textbf{92.29}  &  \textbf{93.30}  &  \textbf{94.71}  &  \textbf{96.12}\\
\cline{1-11}ConfMatrix   	& 64.71   & 62.36   & 72.75   & 78.36  &  83.05  &  87.18  &  90.12  &  92.25  &  93.50  &  95.00\\
\cline{1-11}DARE   			& 64.38   & 66.47   & 76.71   & 81.63  &  84.72  &  86.58  &  88.62  &  90.10  &  91.06  &  91.25\\
\cline{1-11}Entropy   		& 64.71   & \textbf{73.10}   & \textbf{81.61}   & \textbf{86.87}  &  \textbf{89.46}  &  \textbf{91.05}  &  91.59  &  91.77  &  91.90  &  91.92\\
\cline{1-11}Uncertainty   	& 64.71   & \textbf{73.02}   & \textbf{81.33}   & \textbf{86.93}  &  \textbf{89.02}  &  \textbf{90.47}  &  91.24  &  91.69  &  91.77  &  91.92\\
\cline{1-11}AccsExpEntRed   & 63.21   & \textbf{73.23}   & \textbf{80.87}   & \textbf{87.59}  &  \textbf{90.05}  &  \textbf{91.73}  &  \textbf{93.37}  &  \textbf{94.96}  &  \textbf{95.77}  &  \textbf{96.12}\\
\cline{1-11}AccUnc   		& 64.71   & \textbf{72.70}   & \textbf{79.73}   & \textbf{86.55}  &  \textbf{89.46}  &  \textbf{91.64}  &  \textbf{93.30}  &  \textbf{94.80}  &  \textbf{95.82}  &  \textbf{96.12}\\
\cline{1-11}ExpAccUnc  		& 64.57   & \textbf{73.07}   & \textbf{81.21}   & \textbf{87.05}  &  \textbf{89.57}  &  \textbf{91.50}  &  \textbf{93.09}  &  \textbf{94.50}  &  \textbf{95.80}  &  \textbf{96.12}\\
\cline{1-11}
\end{tabular} 
\caption{Results of 3-class-RTE dataset. Rows indicate implemented methods and columns indicate the average number of labels that are used for each problem.}
\label{rteResults3}
\end{table}
\begin{table}
\centering
\begin{tabular}{ccccccccccc}
\textbf{Labels/Sample} & \textbf{1} & \textbf{2} & \textbf{3} & \textbf{4} & \textbf{5} & \textbf{6} & \textbf{7} & \textbf{8} & \textbf{9} & \textbf{10} \\ 
\cline{1-11}MV  & 68.33  & 60.51  & 81.09  & 86.03  & 88.87  & 91.97  & 94.55  & 95.89  & 96.90  & \textbf{97.94}\\
\cline{1-11}Accuracy  & 68.33  & 74.26  & 84.91  & 90.30  & 93.48  & 95.67  & 96.75  & \textbf{97.90}  & \textbf{98.27}  & \textbf{98.70}\\
\cline{1-11}ConfMatrix  & 68.33  & 70.45  & 82.58  & 88.55  & 92.53  & 94.98  & 96.52  & \textbf{97.77}  & \textbf{98.25}  & \textbf{98.29}\\
\cline{1-11}DARE  & 68.33  & 74.03  & 84.22  & 88.16  & 89.61  & 91.19  & 92.53  & 93.46  & 93.03  & 91.13\\
\cline{1-11}Entropy  & 68.33  & \textbf{80.38}  & \textbf{91.27}  & \textbf{95.83}  & \textbf{97.31}  & \textbf{97.65}  & \textbf{97.84}  & \textbf{97.95}  & \textbf{97.97}  & \textbf{97.94}\\
\cline{1-11}Uncertainty  & 68.33  & \textbf{81.15}  & \textbf{91.41}  & \textbf{95.51}  & \textbf{96.65}  & \textbf{97.18}  & \textbf{97.55}  & \textbf{97.84}  & \textbf{97.89}  & \textbf{97.94}\\
\cline{1-11}AccsExpEntRed  & 67.25  & 78.90  & 88.92  & 93.85  & 95.26  & \textbf{96.34}  & \textbf{97.40}  & \textbf{98.14}  & \textbf{98.57}  & \textbf{98.70}\\
\cline{1-11}AccUnc  & 68.33  & \textbf{80.67}  & 89.07  & 93.48  & 95.65  & \textbf{96.45}  & \textbf{97.62}  & \textbf{98.14}  & \textbf{98.59}  & \textbf{98.70}\\
\cline{1-11}ExpAccUnc  & 67.86  & \textbf{80.63}  & 89.52  & 93.83  & 95.52  & \textbf{96.23}  & \textbf{97.12}  & \textbf{98.10}  & \textbf{98.57}  & \textbf{98.70}\\
\cline{1-11}
\end{tabular} 
\caption{Results of 4-class-TEMP dataset. Rows indicate implemented methods and columns indicate the average number of labels that are used for each problem.}
\label{TempResults4}
\end{table}

\textit{One-shot methods.} TEMP is the easiest dataset. According to our expectations, the simpler methods lead to good results. Experiments show that the methods that only model user capabilities lead to best results, but DARE which models the problem difficulty levels has poor performance (Table \ref{TempResults4}). 
In RTE we observed almost the same results as the TEMP dataset (Table \ref{rteResults3} and Fig. \ref{AllMethods_rte}). Since Duchenne is the most challenging dataset and it contains ambiguous problems, we expect to have best results using methods that model both user capabilities and problem difficulty levels. The results meet our expectations (Table \ref{DuResults}).

\begin{figure}
\centering
\includegraphics[scale=0.4]{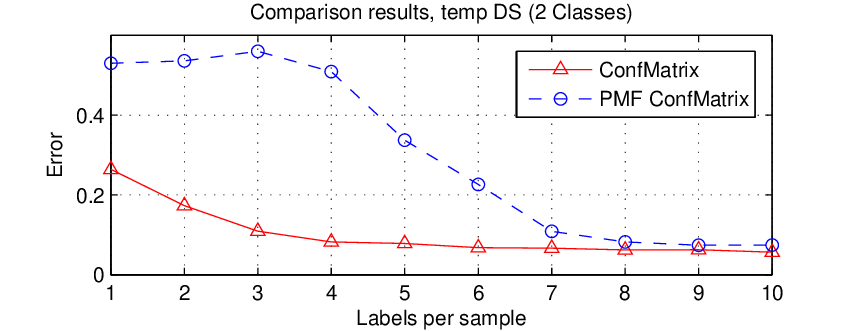}
\caption{The role of matrix completion using PMF prior to label aggregation.}
\label{PMF_Temp}
\end{figure}

\textit{PMF based methods.} The role of matrix completion (using PMF) prior to label aggregation is shown in Table \ref{DuResults} and is illustrated in Fig. \ref{PMF_Temp}. PMF estimates a sparse matrix using a low rank complete matrix. Since all used datasets form almost full-rank data matrices, we don't expect significant results from these methods. Experimental results are compatible with our expectations. In addition, our experiments show that PMF re-estimate the known values well, but it is not able to estimate unseen labels, perfectly. The main reason is that PMF uses mean squared error to measure the distance between the original and the estimated matrices, which is unsuitable for categorical data. 

\begin{figure}
\centering
\includegraphics[scale=0.35]{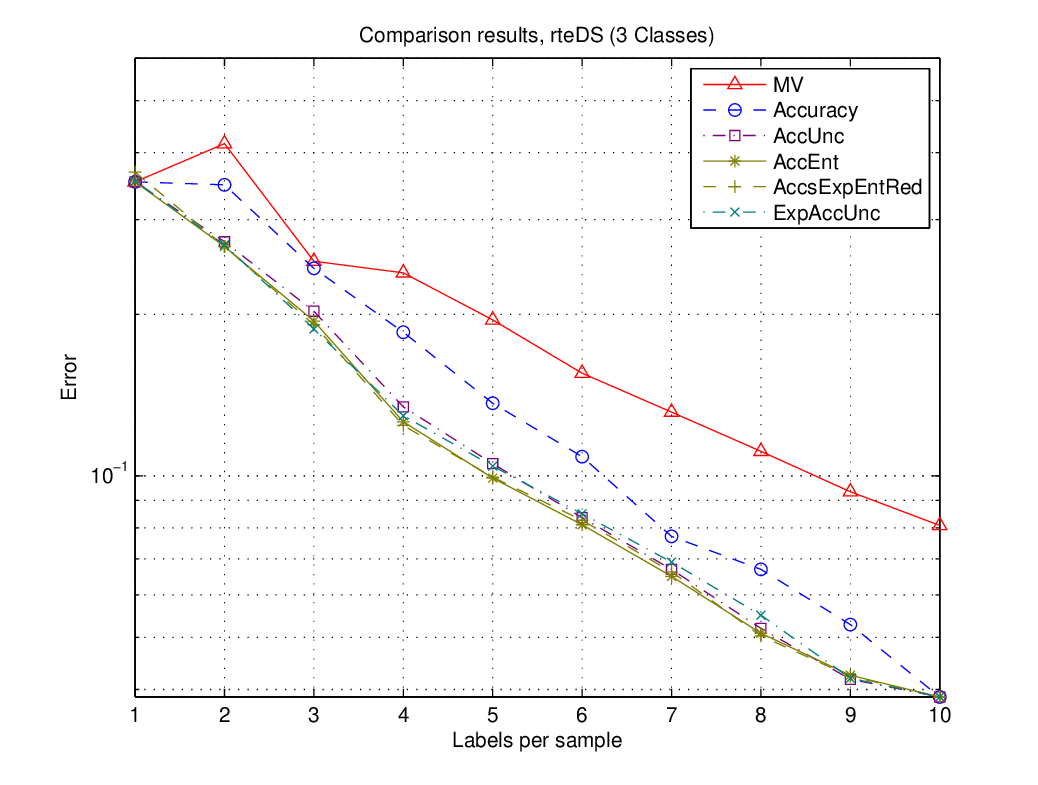}
\caption{Comparison of all accuracy-based (one shot and adaptive) methods on 3-class-RTE dataset.}
\label{AllAccs}
\end{figure}

\textbf{Adaptive methods.} The results show that almost all methods lead to good results in case of having sufficient crowd labels. Adaptive methods lead to better results than the one-shot methods, when a few number of labels per sample are available. In TEMP, the easiest dataset, entropy and uncertainty criteria lead to best results, while the other more complex criteria didn't add any value to these simple ones (Table \ref{TempResults4}). In RTE, the criteria that considers the future, had better results than the simple criteria  (Table \ref{rteResults3} and Fig. \ref{AllAccs}). And, in Duchenne that has most adversarial users, as we expected AccUnc leads to best results, while entropy and uncertainty had poor performances (Table \ref{DuResults}).

\section{Conclusions}

In this paper, we addressed crowd labeling and surveyed the literature using a unified probabilistic framework. Crowd labeling is a popular and growing field which concentrates on solving labeling problems using human crowds. Although this approach has revealed significant performance in different applications, there is no survey associated with its technical aspects. This paper attempts to introduce the challenges associated with this approach and investigates existing methods to overcome these challenges. Since most of these methods model crowd labeling as a statistical inference problem, we proposed a general probabilistic model and showed that all surveyed methods can be considered as a special case. We then described inference algorithms on these models. Sample selection criteria are  introduced for the case when crowd labels are collected adaptively. In addition, the applications suited to the methods and the relationship among them has been studied. Finally, we compared crowd labeling methods based on their merits and demerits and confirmed our analysis using extensive experiments. The surveyed methods are summarized in Table \ref{summary} along with their parameters and inference algorithms.

\begin{table}
\centering
\begin{tabular}{cccc}
 \textbf{Method} & \textbf{Model Parameters} & \textbf{Inference Algorithm} \\
\cline{1-3}
\cite{EM1979} & $\boldsymbol{\pi^j}$ & EM (Likelihood)\\
\cline{1-3}
\cite{txteagle} & $\alpha_j$ & EM (Likelihood) \\
\cline{1-3}
\cite{GLADPaper} & $r_j$ and $1/\beta_j$ & EM (Joint Prob. )\\
\cline{1-3}
\cite{CrowLearn} & $\boldsymbol{\pi^j}$ & EM (Joint Prob. )\\
\cline{1-3}
\cite{MDimWCrowd} & $\{\boldsymbol{\alpha_j}, \gamma_j, \tau_j\}$ and $\boldsymbol{\beta_i}$ & EM (Joint Prob. )\\
\cline{1-3}
\cite{LFM} & $\{\boldsymbol{\alpha_j}, \gamma_j\}$ and $\boldsymbol{\beta_i}$ & EM (Likelihood)\\
\cline{1-3}
\cite{SVD1} & $\alpha_j$ & BP-Style \\
\cline{1-3}
\cite{SVD2} & $\alpha_j$ & SVD Based\\
\cline{1-3}
\cite{VarInf} & $\alpha_j, \boldsymbol{\pi^j}$ & EM/ MF/ BP\\
\cline{1-3}
\cite{PMFC1} & $\alpha$ & PMF + MV\\
\cline{1-3}
\cite{PMFC2} & $\boldsymbol{\pi^j}$ & PMF + EM (Likelihood)\\
\cline{1-3}
\cite{DARE} & $r_j$ and $\{d_i, \delta_i\}$ & EP\\
\cline{1-3} 
\end{tabular} 
\caption{Summary of major surveyed approaches. }
\label{summary}
\end{table}

Several directions of future research are possible. Most of the current crowd labeling models are designed to be used in a wide range of applications and hence they need to be very complex. However, these methods make many simplifications for tractability which are not true in many cases. There are two approaches to overcome this challenge: relaxing the simplifying assumptions and designing task-driven models. There are many aspects in crowd labeling which should be considered in designing a model. For example, most of the existing methods can not be applied to multi-class problems \cite{GLADPaper,SVD1}. Moreover, the correlation among users are usually ignored \cite{Dep2,Dep1,CrowLearn}. In order to overcome these issues in general, more complex models are required. However, complex models need much more labels to estimate the true labels which is not cost efficient. Also, analyzing these methods and providing error bounds is not possible in most of these methods. Therefore, it seems reasonable to switch to application-specific models \cite{LFM,MDimWCrowd} which can use the problem domain properties to simplify the model without non-realistic assumptions.  

There are various applications that can benefit from crowd labeling, but are not compatible with the configuration of current methods. For example, there are different online classification tasks that can utilize crowd labeling to reduce their costs. However, these methods need a pool of available labelers that can provide labels for the incoming data. To the best of our knowledge, there are no marketplaces that support this kind of task and there are very limited number of methods that are designed for online classification using crowd labeling \cite{ahOnline}. Online label acquisition, user expertise modeling in different categories of problems, and handling concept drift in data streams using crowd labeling are samples of requirements of new applications such as market prediction and online advertisement \cite{ahOAI}.

Connecting crowd labeling with related theoretical fields is necessary. For example, since collected crowd labels can be presented as a sparse matrix, working on theories and methods of sparse representation for application on crowd labeling may be a fruitful topic for future work. Probabilistic Matrix Factorization (PMF) has recently been used to estimate unseen crowd labels \cite{PMFC1,PMFC2}. We note that using PMF for crowd labeling requires further work. For example, as our experimental results showed, PMF leads to poor results, because it has its roots in collaborative filtering (CF) which innately differs from crowd labeling; that is,  it deals with ranking data and there are strong correlations between the rows and also the columns of the data matrix in CF. Here, we provide some guidelines to adjust matrix factorization for crowd labeling. Since the data matrix of real crowd datasets is almost full-rank and eliminating the wrong labels reduces the rank of the matrix to $1$, outlier detection is a crucial element to consider before factorization. As a solution, outlier detection can be considered as a term in the objective function of the factorization problem \cite{AOP}. Moreover, a proper distance function must be used in the objective function instead of sum-squared distance, which is not suitable for categorical data. 

\bibliographystyle{agsm}

\section*{Author Biographies}
\leavevmode
\vbox{
\begin{wrapfigure}{l}{80pt}
{\vspace*{-10pt}\fbox{\includegraphics[width=60pt, keepaspectratio]{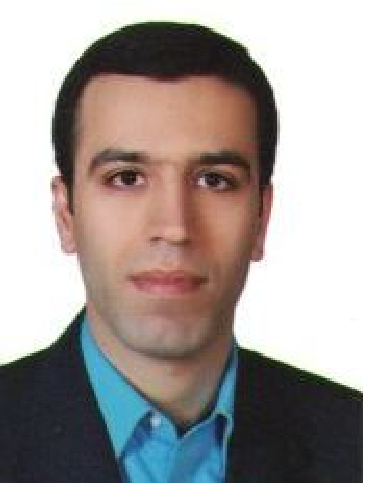}}\vspace*{5pt}}%
\end{wrapfigure}
\noindent\small 
{\bf Jafar Muhammadi} received his M. Sc. degree in Artificial Intelligence from Sharif University of Technology, Tehran, Iran, in 2006. His areas of interest in research include crowdsourcing, human computation, pattern recognition, data mining, and philosophy of mind. He is currently a PhD candidate in the Department of Computer Engineering, Sharif University of Technology, and works as a research assistant at Digital Media Laboratory (DML). The PhD Thesis he has opted for focuses on sample selection in crowd computing. \vadjust{\vspace{40pt}}}
\vbox{
\begin{wrapfigure}{l}{80pt}
{\vspace*{-10pt}\fbox{\includegraphics[width=60pt, keepaspectratio]{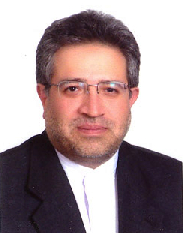}}\vspace*{10pt}}%
\end{wrapfigure}
\noindent\small
{\bf Hamid R. Rabiee} received his BS and MS degrees in Electrical Engineering from CSULB, (with great distinction), his EEE degree in Electrical and Computer Engineering from USC, and his PhD in Electrical and Computer Engineering from Purdue University, West Lafayette, in 1996. From 1993 to 1996, he was a Member of Technical Staff at AT\&T Bell Laboratories. From 1996 to 1999, he worked as a Senior Software Engineer at Intel Corporation. He was also with PSU, OGI and OSU Universities as an Adjunct Professor of Electrical and Computer Engineering from 1996-2000. Since September 2000, he has joined Sharif University of Technology, Tehran, Iran. He is the founder of Sharif University Advanced Information and Communication Technology Research Center (AICT), Sharif University Advanced Technologies Incubator (SATI), Sharif Digital Media Laboratory (DML), and Sharif Mobile Value Added Services Laboratory (VASL). He is currently a Professor of Computer Engineering at Sharif University of Technology, and the Director of AICT, DML, and VASL. He has been the initiator and director of national and international level projects in the context of National ICT Development Plan and UNDP International Open Source Network (IOSN). He has received numerous awards and honors for his industrial, scientific and academic contributions. He has acted as chairman in a number of national and international conferences, and holds three patents. He is also a Senior Member of IEEE. \vadjust{\vspace{20pt}}}
\vbox{
\begin{wrapfigure}{l}{80pt}
{\vspace*{-10pt}\fbox{\includegraphics[width=60pt, keepaspectratio]{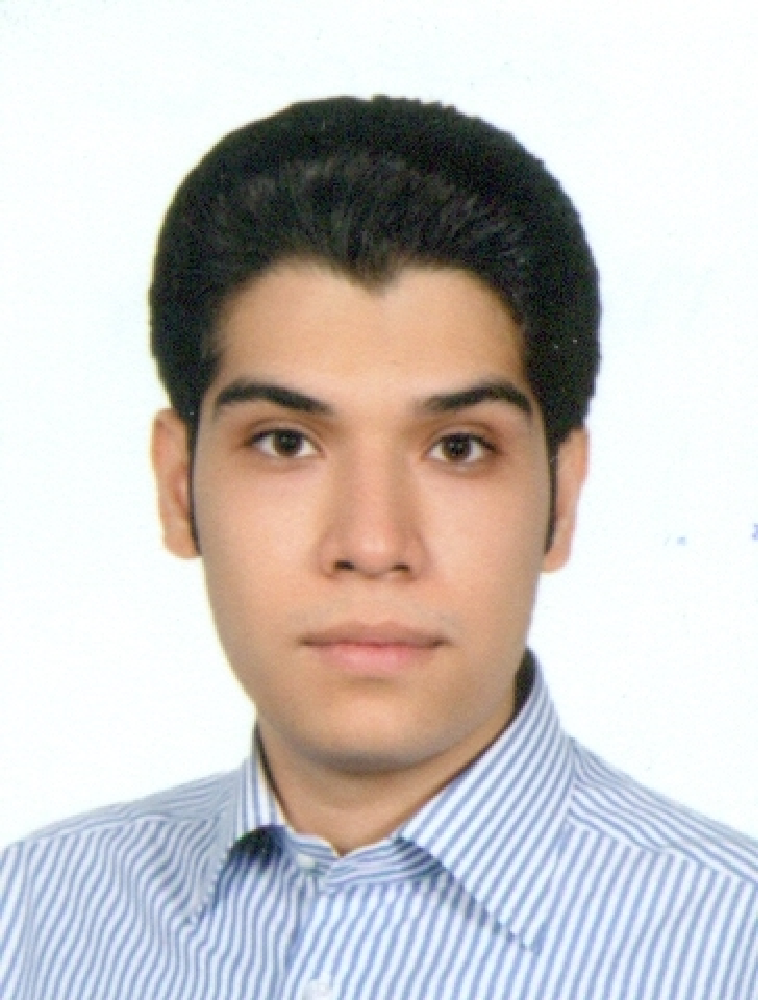}}\vspace*{10pt}}%
\end{wrapfigure}
\noindent\small 
{\bf Abbas Hosseini} received his B.Sc. in Software Engineering from Sharif University of Technology, Tehran, Iran, in 2012 and is currently working towards his M.Sc. degree in the Department of Computer Engineering at Sharif University of Technology. His current research interests include probabilistic graphical models and Bayesian non-parametric methods and their application to large scale data mining.\vadjust{\vspace{20pt}}}
\correspond{Hamid R. Rabiee, Department of Computer Engineering, Sharif University of Technology, Tehran, Iran. Email: rabiee@sharif. edu}
\label{lastpage}
\end{document}